\documentclass[runningheads]{llncs}
\usepackage{graphicx}

\begin{document}

\title{Enhancing Knee Osteoarthritis severity level classification using diffusion augmented images}

\titlerunning{Knee Osteoarthritis severity level classification}




\author{Paleti Nikhil Chowdary\inst{1}\orcidID{0009-0002-2300-0997} \and
Gorantla V N S L Vishnu Vardhan\inst{1}\orcidID{0009-0007-1821-5586} \and
Menta Sai Akshay\inst{1}\orcidID{0009-0007-3130-4209} \and
Menta Sai Aashish\inst{1}\orcidID{0009-0008-5189-1882} \and
Vadlapudi Sai Aravind\inst{1}\orcidID{0009-0002-8692-3860} \and
Garapati Venkata Krishna Rayalu\inst{1}\orcidID{0009-0007-5971-0743} \and
Aswathy P \inst{1}\orcidID{0000-0002-8641-1139}
}

\institute{
Amrita School of Artificial Intelligence, Coimbatore, \\ Amrita Vishwa Vidyapeetham, India. \\
\email{nikhil.paleti@outlook.com},
\email{vishnuvardhangorantla0308@gmail.com},
\email{akshaymenta24@gmail.com},
\email{aashishmenta249@gmail.com},
\email{aravindvadlapudi2003@gmail.com},
\email{gkrishnarayalu@gmail.com},
\email{p\_aswathy@cb.amrita.edu}
}

\authorrunning{Paleti Nikhil Chowdary et al.}

\maketitle              
\begin{abstract}
This research paper explores the classification of knee osteoarthritis (OA) severity levels using advanced computer vision models and augmentation techniques. The study investigates the effectiveness of data preprocessing, including Contrast-Limited Adaptive Histogram Equalization (CLAHE), and data augmentation using diffusion models. Three experiments were conducted: training models on the original dataset, training models on the preprocessed dataset, and training models on the augmented dataset. The results show that data preprocessing and augmentation significantly improve the accuracy of the models. The EfficientNetB3 model achieved the highest accuracy of 84\% on the augmented dataset. Additionally, attention visualization techniques, such as Grad-CAM, are utilized to provide detailed attention maps, enhancing the understanding and trustworthiness of the models. These findings highlight the potential of combining advanced models with augmented data and attention visualization for accurate knee OA severity classification.

\end{abstract}
\section{Introduction}

Osteoarthritis (OA) is a chronic degenerative condition marked by cartilage degradation that worsens over time and finally results in bone deterioration. Knee osteoarthritis (KOA), one of its many variations, primarily affects The medial, lateral, and patellofemoral joints—the three compartments of the knee joint. It usually develops gradually over a period of 10 to 15 years, causing interruptions in daily life \cite{ref1,ref2}.

According to Wang et.al \cite{ref8}, in patients over 60 years of age, the prevalence of symptomatic KOA ranges from 10.0\% to 16.0\%, whereas that of radiographic KOA ranges from 35.0\% to 50.0\%. The prevalence of KOA has increased by twofold in men and treble in women in the USA during the past 20 years, affecting around 250 million individuals globally placing a heavy burden on society.

The diagnosis of KOA typically relies on symptom evaluation, arthroscopy, X-rays, and Magnetic Resonance Imaging (MRI). The Kellgren and Lawrence grading system presented by Kellgren JH in \cite{ref9} is the most commonly used radiographic classification for knee osteoarthritis. The osteoarthritis grading system includes uncertain Joint space narrowing and potential osteophytic lipping (Grade 1), osteophytes with clear signs of joint space narrowing, sclerosis, and potential bony deformity (Grade 2), multiple osteophytes with clear signs of joint space narrowing, sclerosis, and potential bony deformity (Grade 3), and prominent osteophytes with clear signs of joint space narrowing, severe sclerosis, and certain bony deformity (Grade 4).

Though there have been various studies tackling the classification of KOA, there hasn't been a study on using state-of-the-art augmentation methods like diffusion models and very few works presented the reasons for model prediction through explainable AI. In this study, our aim is to effectively classify knee osteoarthritis X-ray images into 5 classes and also present the effectiveness of using data preprocessing and data augmentation. We will also explore the reasons behind the predictions of the classification models by analyzing the gradcam images which provide an explanation for the classifications done by the model.

Three different experiments were conducted, In the first experiment various computer vision models were trained using the original OAI dataset, In the second one training was done using preprocessed OAI dataset which was constructed based on CLAHE \cite{ref10} method. In the last experiment, we generated synthetic data using diffusion models to augment the dataset and trained the computer vision models on the augmented dataset. Finally, a comparison study was done among the models and finally, trained model gradients were utilized to plot gradcam images which showed the relevant parts of the image responsible for classification.

The subsequent section will provide a summarization of the relevant literature on knee osteoarthritis, followed by an explanation of the methodology employed in this study. Subsequently, the results and discussion section will present the findings derived from our experiments and delve into their analysis. Finally, the concluding section will present our overall conclusion.

\section{Related Works}

In \cite{ref4}, Fabi Prezja et al. used data augmentation with a WGAN to generate synthetic X-ray DeepFake images for knee osteoarthritis (KOA). Surgeons achieved 65.28\% accuracy and radiologists achieved 59.40\% accuracy in classifying real or fake images. In \cite{ref6}, Joseph Humberto Cueva et al. developed a semi-automatic CADx model using Deep Siamese CNNs and ResNet-34. Their model achieved a 61\% accuracy for KOA detection and classification. In \cite{ref7}, Kevin A Thomas et al. used traditional augmentation techniques and a custom CNN model with ImageNet weights to automate KOA severity classification. They achieved 71\% accuracy and an f1-score of 0.70. In \cite{ref5}, Abdelbasset Brahima et al. created a decision support tool for early KOA detection using X-ray imaging and machine learning. They achieved an accuracy of 82.90\% with a random forest classifier.

These studies highlight various methods for processing KOA datasets, including traditional techniques and advanced approaches like GANs. However, there is limited research on using diffusion models for data augmentation and integrating them with advanced traditional processing techniques. This research aims to investigate the effectiveness of diffusion models for data augmentation and the impact of advanced image-processing techniques on KOA classification.

\section{Methodology}

\begin{figure}
    \centering
    \includegraphics[width=250pt]{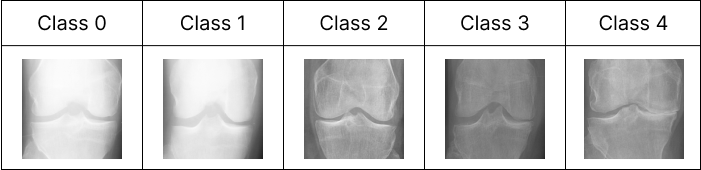}
    \caption{Preprocessed Images}
    \label{fig:preprocessed}
\end{figure}

We first preprocess the dataset using Contrast-Limited Adaptive Histogram Equalization (CLAHE) technique and prepared a preprocessed dataset which is used to train Diffusion models. Figure \ref{fig:preprocessed} displays the preprocessed Images for each class. Images are then generated from the diffusion models using a  Denoising Diffusion Implicit Models (DDIM) scheduler and a final dataset is prepared by combining all the images. then various experiments are run as stated before and the current section describes all the aspects in detail.

\subsection{Dataset}

The OAI dataset encompasses a comprehensive collection of X-ray images focusing on knee joint detection and knee KL grading. Knee Osteoarthritis Severity Grading Dataset \cite{ref11} is organized from the OAI dataset which encompasses 9786 knee X-ray images of size 224x224x1, categorized into five distinct groups, known as KL-grades, which provide an indication of the severity of knee osteoarthritis. Detailed information about the dataset is presented Table \ref{org_dat}. 

\begin{table}
\begin{center}
\caption{OAI Dataset}\label{org_dat}
\begin{tabular}{|c|c|c|c|c|}
\hline
\textbf{Severity (Class)} & \textbf{Train (75\%)} & \textbf{Test (15\%)} & \textbf{Valid (10\%)} & \textbf{Total}\\
\hline
0 & 2,890 & 639 & 328 & 3,857\\
1 & 1,321 & 296 & 153 & 1,770\\
2 & 1,919 & 447 & 212 & 2,578\\
3 & 957 & 223 & 106 & 1,286\\
4 & 217 & 51 & 27 & 295\\
\hline
\end{tabular}
\end{center}
\end{table}

In this study, data preprocessing was performed using CLAHE \cite{ref10} to enhance the quality and usability of the dataset. 
The block size was set to 8x8 pixels and the clip limit to 0.03 for preprocessing.



\subsection{Data Augmentation}


In order to address a dataset's limited amount of image samples and reduce the possibility of bias toward particular classes, data augmentation is essential. Diffusion models are one effective method for data augmentation. Diffusion models develop an invertible generative process that enables the creation of fresh data samples based on the discovered distribution. We can add artificial images to our dataset that accurately reflect the statistical characteristics of the original data by utilizing diffusion models.

\begin{figure}
    \centering
    \includegraphics[width=250pt]{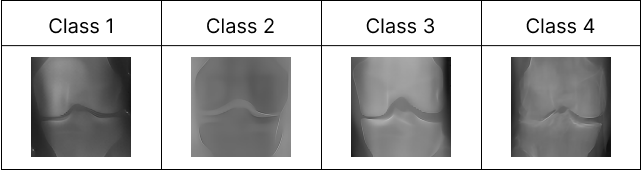}
    \caption{Generated Images for each Class}
    \label{fig:augmented}
\end{figure}

As the preferred diffusion model for data augmentation in our study, we used denoising diffusion implicit models (DDIMs). The pre-processed images were initially scaled to (64 x 64 x 1) dimensions before DDIMs were trained on them. Then, 200 photos for each class except class 0 were produced using the trained DDIM. Figure \ref{fig:augmented} shows a sample-generated image from each class


Diffusion models for data augmentation, like DDIMs, offer a potent method to increase the dataset's diversity and enhance the generalization skills of the trained models. Additionally, it is important to note that a DDIM scheduler, which chooses the level of diffusion at each training, was in charge of managing and directing the augmentation process step. This scheduler aids in regulating the trade-off between the diversity of the generated samples and their resemblance to the original data, so achieving a balance that improves model performance and reduces potential bias.

The original dataset consists of knee osteoarthritis images with dimensions of 224 x 224. To improve visualization and analysis, we apply the CLAHE technique for image preprocessing. However, considering computational constraints, we resize the preprocessed images to 64 x 64 for training the diffusion model.

\begin{figure}
    \centering
    \includegraphics[width=250pt]{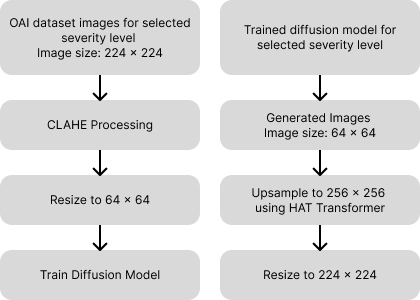}
    \caption{Diffusion Methodology}
    \label{fig:Diffusion}
\end{figure}

To augment the dataset, we employ a trained diffusion model to generate additional images. For each class, except the control class, we generate 100 augmented images. Since the desired image size for classification is 224 x 224, we upscale the augmented images (originally 64 x 64) using the HAT Transformer and chaiNNer GUI. This upscale process increases the dimensions to 256 x 256, and we subsequently resize the images back to 224 x 224 using the Lancos interpolation method, which is clearly explained in Figure \ref{fig:Diffusion}

\subsection{Training}
\begin{figure}
    \centering
    \includegraphics[width=250pt]{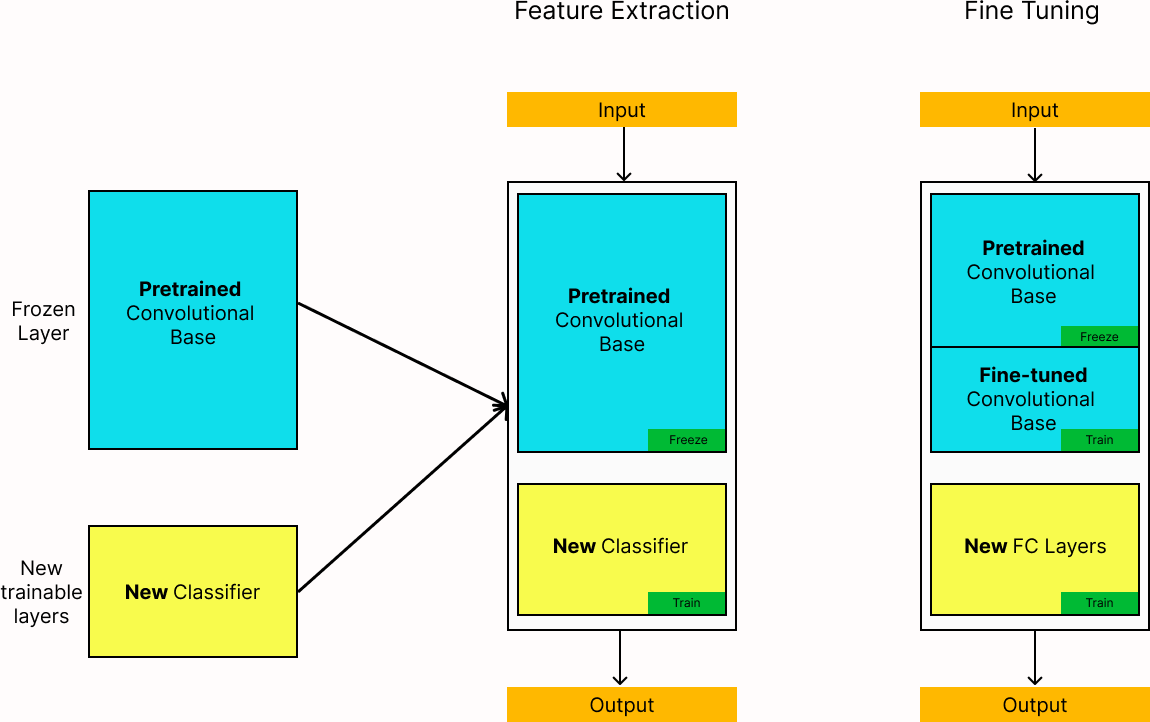}
    \caption{Feature Extraction Vs Fine-tuning}
    \label{fig:training}
\end{figure}

In this work, we examined feature extraction and fine-tuning as two methods for training vision models. Both methods made use of an augmented dataset, but we chose to concentrate on the benefits of fine-tuning rather than feature extraction. Figure \ref{fig:training} illustrates the visual differences in feature extraction and fine-tuning processes.

We make sure that the pre-trained weights, which capture important knowledge from a different task or domain, are kept in the fine-tuning procedure by initially freezing the basic model and training the models for a specific number of epochs. As a result, the model can take advantage of the underlying model's extensive representations and feature extraction skills.

We unfroze a portion of the base model layers after the initial training phase, usually the final 15 layers, and we continued training for several epochs. This action permits the pretrained weights are adjusted by the model to the particular task at hand, fine-tuning them to the subtleties and specifics of the augmented dataset. The model can learn task-specific attributes necessary for optimum performance on the target task by updating these unfrozen layers.

After preparing the datasets, The model was optimised using Lora with a projection rank of 16 after being loaded into memory with 8-bit precision. The optimised model was then trained for 10 epochs with an 8-batch size and a ${10^{ - 3}}$ learning rate.

\section{Results and Discussion}
The results obtained from the experiments are summarized in Table \ref{Comparison of various models on 3 types of dataset}. The table presents the accuracy scores of different models when trained on three different datasets: original, preprocessed, and augmented. These results provide insights into the performance of the models and the impact of data preprocessing and augmentation techniques on the classification of knee osteoarthritis (OA) severity levels.

\begin{table}
\begin{center}
\caption{Comparison of various models on 3 types of dataset}
\label{Comparison of various models on 3 types of dataset}
\begin{tabular}{|c|c|c|c|}
\hline
\textbf{Model} & \textbf{Original} & \textbf{Preprocessed} & \textbf{Augmented} \\
\hline
Xception & 52\% & 76\% & 79\%\\
VGG16 & 78\% & 77\% & 82\%\\
ConvNeXtTiny & 72\% & 80\% & 81\%\\
EfficientNet B3 & 68\% & 76\% & 84\%\\
Densenet 201 & 60\% & 76\% & 79\%\\
Vision Transformer & 69\% & 71\% & 72\% \\
Swin Transformer V2 & 72\% & 73\% & 73\% \\
\hline
\end{tabular}
\end{center}
\end{table}

Overall, the results indicate that data preprocessing and augmentation techniques have a positive impact on the performance of the models for classifying knee OA severity levels. Data augmentation, in particular, consistently improved the accuracy of the models across different architectures. The EfficientNetB3 model achieved the highest accuracy of 84\% on the augmented dataset, showcasing the potential of combining advanced models with augmented data for accurate classification of knee OA severity.

\subsection{Visualizing GradCam}

In this work, we discuss the critical requirement for visualising attention in convolutional neural networks (CNNs) and vision transformers (ViTs) in this research. In order to comprehend how these models pay attention to various areas of an image and make decisions, attention visualisation approaches are essential. We specifically look at Grad-CAM, CAM (Class Activation Mapping), and self-attention heatmaps as popular attention visualisation techniques.

\begin{figure}
    \centering
    \includegraphics[width=300pt]{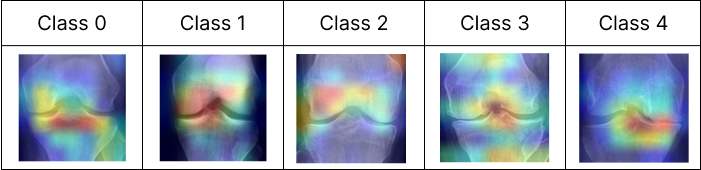}
    \caption{AttentionMap Visualization Using Gradcam}
    \label{fig:Gradcam}
\end{figure}

Grad-CAM, which can offer precise attention maps for intermediate layers as shown in Figure \ref{fig:Gradcam}, is our favoured technology among these ones. Grad-CAM can be used with different CNN and ViT designs because it is model-agnostic. Grad-CAM has the unique ability to overlay attention maps on the original image, providing a clear visual representation of the model's attentional focus.

In this study, we use the EfficientNetB3 model to classify osteoarthritis, paying special attention to the spacing between joints as a crucial characteristic. This agreement with osteoarthritis severity indicators improves our comprehension of the model's decision-making process. We seek to enhance the interpretability, dependability, and trustworthiness of CNNs and ViTs by visualizing attention using methods such as Grad-CAM. The quality and fairness of these models are improved overall by using attention visualization approaches, which also help in locating biases or flaws in the model.




\section{Conclusion and Future Scope}

In Conclusion, Convolutional Neural Network (CNN) models are effective in correctly classifying OA severity levels, as shown by our study on the classification of knee osteoarthritis (OA) severity levels. We have identified the most efficient methods for pre-processing and data augmentation through the investigation of diffusion-based augmentation, explainable AI (Grad-CAM), and a comparison of different CNN and transformer designs. The Efficientnet-B3 performed best among the models tested, displaying greater accuracy in OA severity classification. Through automated analysis of knee X-ray images, this research shows potential for bettering the diagnosis and treatment of knee OA, facilitating informed choice-making, and boosting patient care. The results highlight the potential of deep learning techniques to solve the difficulties involved in assessing knee OA, opening the door for more specialized treatment strategies and improved patient outcomes.

Investigating various pre-processing techniques and contrasting their effects on model performance will be part of future work in the categorization of knee osteoarthritis. Finding the best strategy for enhancing model accuracy and generalization will also depend on researching and evaluating the efficacy of various augmentation strategies. 

\bibliographystyle{splncs04} 

\bibliography{refs.bib}

\begin{thebibliography}{10}
\providecommand{\url}[1]{\texttt{#1}}
\providecommand{\urlprefix}{URL }
\providecommand{\doi}[1]{https://doi.org/#1}

\bibitem{ref4}
Brahim, A., Jennane, R., RIAD, R., Janvier, T., Khedher, L., Toumi, H.,
  Lespessailles, E.: A decision support tool for early detection of knee
  osteoarthritis using x-ray imaging and machine learning: Data from the
  osteoarthritis initiative. Computerized Medical Imaging and Graphics
  \textbf{73} (04 2019). \doi{10.1016/j.compmedimag.2019.01.007}

\bibitem{ref11}
Chen, P.: Knee osteoarthritis severity grading dataset. Mendeley Data
  \textbf{1} (2018). \doi{10.17632/56rmx5bjcr.1}

\bibitem{ref5}
Cueva, J., Castillo, D.P., Espinos~Morato, H., Duran, D., Díaz, P.,
  Lakshminarayanan, V.: Detection and classification of knee osteoarthritis.
  Diagnostics  \textbf{12}, ~2362 (09 2022)

\bibitem{ref9}
Kellgren, J.H., Lawrence, J.S.: Radiological assessment of osteo-arthrosis.
  Annals of the Rheumatic Diseases  \textbf{16}(4),  494--502 (1957).
  \doi{10.1136/ard.16.4.494}, \url{https://ard.bmj.com/content/16/4/494}

\bibitem{ref1}
Lespasio, M.: Knee osteoarthritis: A primer. The Permanente Journal
  \textbf{21} (12 2017). \doi{10.7812/TPP/16-183}

\bibitem{ref10}
Reza, A.: Realization of the contrast limited adaptive histogram equalization
  (clahe) for real-time image enhancement. VLSI Signal Processing  \textbf{38},
   35--44 (08 2004). \doi{10.1023/B:VLSI.0000028532.53893.82}

\bibitem{ref7}
Song, J., Meng, C., Ermon, S.: Denoising diffusion implicit models. CoRR
  \textbf{abs/2010.02502} (2020), \url{https://arxiv.org/abs/2010.02502}

\bibitem{ref6}
Thomas, K., Kidzinski, L., Halilaj, E., Fleming, S., Venkataraman, G., Oei, E.,
  Gold, G., Delp, S.: Automated classification of radiographic knee
  osteoarthritis severity using deep neural networks. Radiology: Artificial
  Intelligence  \textbf{2},  e190065 (03 2020). \doi{10.1148/ryai.2020190065}

\bibitem{ref2}
Veronese, N., Honvo, G., Bruyère, O., Rizzoli, R., Barbagallo, M., Maggi, S.,
  Smith, L., Sabico, S., Al-Daghri, N., Cooper, C., Pegreffi, F., Reginster,
  J.Y.: Knee osteoarthritis and adverse health outcomes: an umbrella review of
  meta-analyses of observational studies. Aging Clinical and Experimental
  Research  \textbf{35} (11 2022). \doi{10.1007/s40520-022-02289-4}

\bibitem{ref8}
Wang, L., Lu, H., Chen, H., Jin, S., Wang, M., Shang, S.: Development of a
  model for predicting the 4-year risk of symptomatic knee osteoarthritis in
  china: a longitudinal cohort study. Arthritis Research \& Therapy
  \textbf{23},  1--13 (2021)

\end{thebibliography}

\end{document}